%% file: main.tex
\documentclass[a4paper, 10pt, conference]{style/ieeeconf}

\IEEEoverridecommandlockouts
\usepackage{graphics} 
\usepackage{epsfig} 
\usepackage{mathptmx} 
\usepackage{times} 
\usepackage{amsmath} 
\usepackage{amssymb}  
\usepackage{siunitx} 
\usepackage[inkscapelatex=false]{svg}

\usepackage[hidelinks]{hyperref}

\usepackage{subcaption} 
\usepackage{pdfpages} 

\usepackage{algorithm}
\usepackage{algpseudocode}

\usepackage{authblk}

\usepackage{calc}
\usepackage{array}
\usepackage{tabulary}

\DeclareMathAlphabet{\mathcal}{OMS}{cmsy}{m}{n}

\usepackage{mathtools}

\title{\LARGE \bf
Reference Free Platform Adaptive Locomotion for Quadrupedal Robots using a Dynamics Conditioned Policy
}

\author[1]{David Rytz\thanks{
This work was supported by the Engineering and Physical Sciences Research Council (EPSRC) through a Doctoral Training Account (DTA) studentship at the University of Oxford. For the purpose of open access, the author has applied a Creative Commons Attribution (CC BY) licence to any Author Accepted Manuscript version arising. (Corresponding author: rytz@robots.ox.ac.uk.)
}}
\author[2]{Suyoung Choi}
\author[1]{Wanming Yu}
\author[1]{Wolfgang Merkt}
\author[2]{Jemin Hwangbo}
\author[1]{Ioannis Havoutis}
\affil[1]{Dynamic Robot Systems, Oxford Robotics Institute, University of Oxford}
\affil[2]{RaiLab, Department of Mechanical Engineering, KAIST}

\begin{document}

\maketitle

\input{sections/abstract}
\input{sections/introduction}
\input{sections/preliminaries}

\input{sections/methodology}

\input{sections/results_and_discussion}

\input{sections/conclusion_and_future_work}

\bibliographystyle{style/IEEEtran}
\bibliography{references.bib}

\end{document}

%% file: sections/abstract.tex
\begin{abstract}
This article presents \textbf{P}latform \textbf{A}daptive \textbf{L}ocomotion (PAL), a unified control method for quadrupedal robots with different morphologies and dynamics. We leverage deep reinforcement learning to train a single locomotion policy on procedurally generated robots. The policy maps proprioceptive robot state information and base velocity commands into desired joint actuation targets, which are conditioned using a latent embedding of the temporally local system dynamics. We explore two conditioning strategies - one using a GRU-based dynamics encoder and another using a morphology-based property estimator - and show that morphology-aware conditioning outperforms temporal dynamics encoding regarding velocity task tracking for our hardware test on ANYmal C. Our results demonstrate that both approaches achieve robust zero-shot transfer across multiple unseen simulated quadrupeds. Furthermore, we demonstrate the need for careful robot reference modelling during training: exposing the policy to a diverse set of robot morphologies and dynamics leads to improved generalization, reducing the velocity tracking error by up to 30\% compared to the baseline method. Despite PAL not surpassing the best-performing reference-free controller in all cases, our analysis uncovers critical design choices and informs improvements to the state of the art. 

\end{abstract}

%% file: sections/introduction.tex
\section{Introduction}

\label{sec:introduction}
The increasing prevalence of quadrupedal robots in research and industry allows for autonomous inspection and surveillance applications and enables their use in search and rescue operations. Recent advances in numerical optimization ~\cite{Bellicoso2018, mastalliCrocoddylEfficientVersatile2020} and Reinforcement Learning (RL)~\cite{hwangboLearningAgileDynamic2019, gangapurwalaRLOCTerrainAwareLegged2022, pengLearningAgileRobotic2020, zhuangRobotParkourLearning2023, jeneltenPerceptiveLocomotionRough2020} based control approaches have made essential contributions to the development of quadrupedal locomotion and navigation in complex environments with impressive agility. However, most existing control designs develop robot-specific controllers, optimized for a single robot system, and require extensive retraining and parameter tuning whenever the robot's morphology or dynamics change.

These platform-specific approaches create a fundamental challenge: developing a controller for each new quadruped requires notable computational resources, laborious manual parameter tuning, and reward engineering. As the diversity of commercial quadrupeds increases, ranging from lightweight (\SI{12}{kg}--e.g., A1~\cite{unitreeroboticsA12023}) to heavy-duty (\SI{50}{kg} ANYmal C~\cite{ackermanevanANYboticsIntroducesSleek2019}) systems, this line of approaches becomes increasingly hard to scale without specific engineering and hardware expertise.

We identify three central challenges: first, different quadrupeds exhibit substantial variations in mass distribution, joint configurations, actuator properties, and kinematic structures. Second, as these variations fundamentally change the locomotion dynamics, they require different gait patterns and control strategies. Third, even seemingly appearing robots can display significant differences in control latency and actuator response, creating a substantial sim-to-real gap when deploying learned controllers.

Model-based approaches~\cite{grandiaContactInvariantModel2018, deisenrothPILCOModelBasedDataEfficient2011} have emerged to tackle the challenge of generating effective locomotion control actions while modelling the system dynamics. In contrast, learning-based approaches~\cite{haWorldModels2018, huangOnePolicyControl2020} demonstrated higher sim-to-real success rate and robustness in quadruped locomotion, and therefore, this work focuses on the latter approach.

\begin{figure*}
    \centering
    \includegraphics[width=0.85\linewidth]{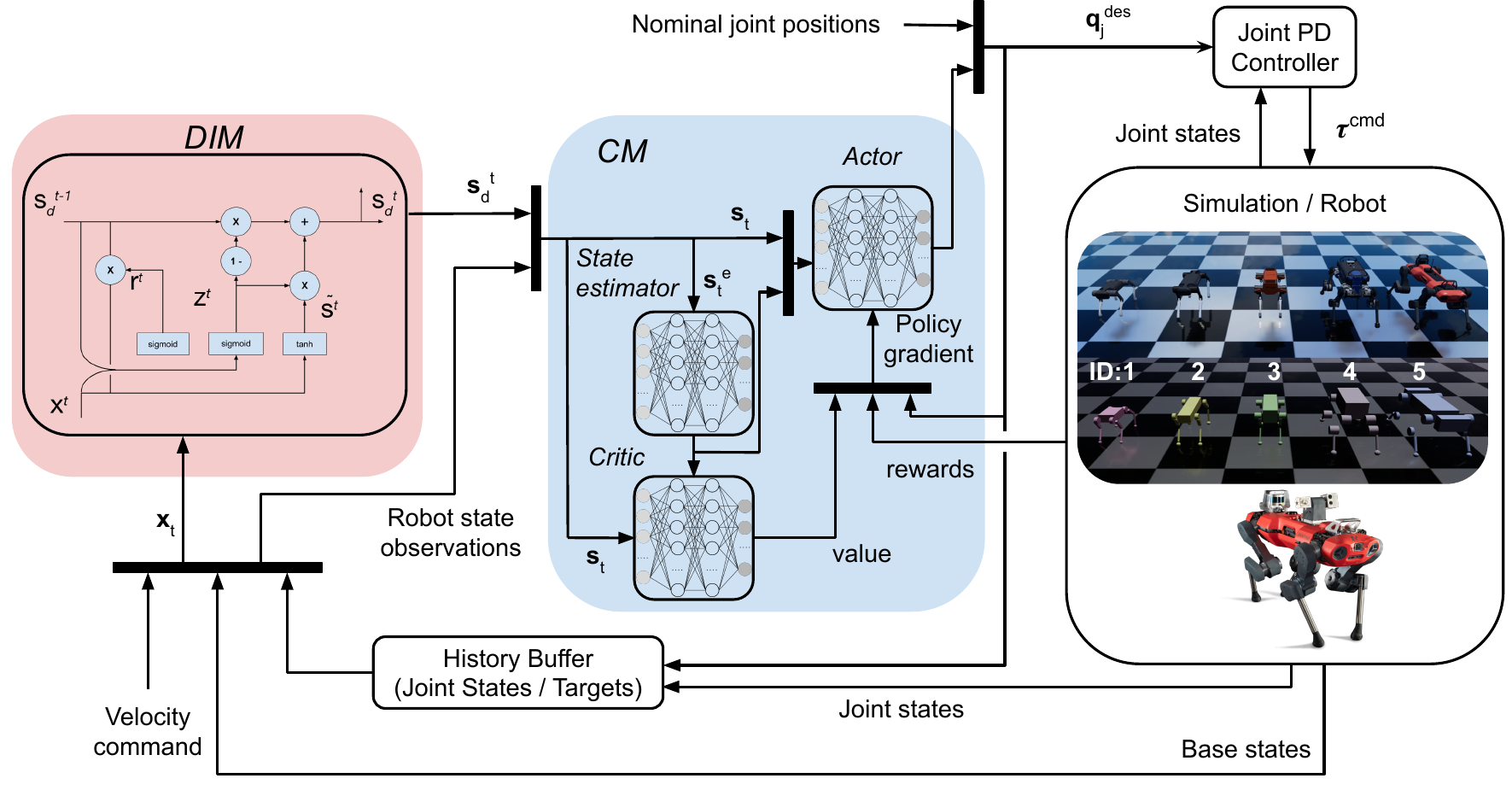}
    \caption{Overall pipeline. A Dynamic Inference Module (DIM) in the form of a gated recurrent unit generates an approximate representation $\mathbf{s}_d$ of the dynamics by observing how the state $\mathbf{x}_t$ changes. Then, the Control Module (CM) utilizes this information, represented by $\mathbf{s}_t$, to generate a control action $\mathbf{q}^{des}$ allowing the quadruped to locomote through the respective joint PD controller. We replace the DIM accordingly for the baseline implementation and adjust the observation state input to the CM actor and critic~\cite{luoMorALLearningMorphologically2024}. The control frequency of \SI{100}{\hertz} for both modules is applied while changing the set of robot IDs during training into subsets of $\left\{1,2,3,4,5\right\}$ ranging from the quadruped A1 to ANYmal C.}
    \label{fig:introduction:control_framework}
\end{figure*}

Imitation learning or using motion priors is a popular approach to finding a \textit{universal} locomotion controller for quadrupeds. \textit{Feng et al.} utilize imitation learning based on motion capture data combined with a large state-action history to train a phase-variable locomotion controller ~\cite{fengGenLocoGeneralizedLocomotion2022}. They successfully deploy their controller on hardware while showing that it works well for systems of about \SI{10}{\kg}. Still, the performance deteriorates for medium-sized quadrupeds larger than \SI{20}{\kg}. This limitation highlights the challenge of scaling controllers to substantially different robot dynamics. \textit{Shafiee et al.} tackle this challenge by training a policy to modulate the parameters of a central pattern generator to enable a wide range of simulated robots to locomote by tracking the resulting trajectories~\cite{shafieeManyQuadrupedsLearningSingle2023}. They managed to deploy it on hardware systems of \SI{12}{kg} and up to \SI{200}{kg} in simulation. However, the reliance on motion templates may constrain behavioral adaptability in complex environments. \textit{Zargarbashi et al.} improve these approaches by reducing the dependence on large state-action history by using a Gated Recurrent Unit (GRU) as a memory storage~\cite{zargarbashiMetaLocoUniversalQuadrupedal2024}. While working on small and medium-sized quadrupeds, their approach still uses reference motions, which may limit emergent behaviors. 

Taking a fundamentally different approach, \textit{Luo et al.} concurrently train a control policy and a morphology prediction network, allowing the controller to discover joint position targets without relying on motion priors, allowing adaptive and challenging terrain locomotion. They successfully implemented it on a range of simulated quadrupeds and the A1 and Go1 hardware with modifications weighing up to \SI{30}{kg} with significant disturbances \cite{luoMorALLearningMorphologically2024}. \textit{Bohlinger et al.} present a single locomotion network, with implicit robot kinematic and dynamic encoder capabilities using attention encoding. They demonstrate promising generalization capabilities in simulation, focusing more on general multi-legged locomotion capabilities and less on improving robustness and closing the sim-to-real gap for quadrupedal locomotion~\cite{bohlingerOnePolicyRun2025}. Further, they only demonstrate hardware deployment to small quadrupeds at \SI{14}{\kg}. 

This paper introduces a novel method to learn a reference-free locomotion policy for quadrupedal robots adaptable to platforms with diverse dynamic and kinematic properties. We take inspiration from \cite{zargarbashiMetaLocoUniversalQuadrupedal2024} using a GRU for the embedding of robot parameters, as depicted in Figure~\ref{fig:introduction:control_framework}. Our controller discovers the joint commands in a reference-free manner as in \cite{luoMorALLearningMorphologically2024}, thus not using motion templates. Our proposed control system integrates the Dynamics Inference Module (DIM), which is pivotal in continuously estimating and improving a latent representation for robots with different kinematic and dynamic properties, including those unseen during training. The Control Module (CM) generates appropriate control actions by leveraging these latent dynamics. We evaluate the effectiveness of our approach by assessing its applicability to other quadrupeds after training the DIM and CM iteratively instead of concurrently, as in \cite{luoMorALLearningMorphologically2024}. Additionally, we incorporate explicit latency modelling into the training process to facilitate deployment on robots with highly non-linear actuator properties~\cite{hwangboLearningAgileDynamic2019}. The absence of motion priors necessitates us to include a base linear velocity estimator modelled after \cite{jiConcurrentTrainingControl2022a}.

While ~\cite{luoMorALLearningMorphologically2024, fengGenLocoGeneralizedLocomotion2022, zargarbashiMetaLocoUniversalQuadrupedal2024} use a single reference robot model and scale its kinematic and dynamic properties, and \cite{shafieeManyQuadrupedsLearningSingle2023} chooses to incorporate a range of robots,  we show how the careful choice of diverse reference robots during training influences the robustness and quality of the resulting policy. This approach enables successful generalization and deployment from the models seen during training, including the \SI{50}{kg} ANYmal C robot -- extending beyond the \SI{30}{kg} limit demonstrated in previous work for zero-shot deployments onto hardware~\cite{luoMorALLearningMorphologically2024}.

Our contributions include: (1) a novel training method including a DIM and a single platform-adaptive RL locomotion policy with a wide range of kinematics and dynamics randomization, (2) comprehensive comparison with state-of-the-art reference-free \textit{universal} locomotion controller, demonstrating improved robustness, (3) successful zero-shot transfer to various quadrupeds in simulation (ranging from the A1 with \SI{12}{\kg} to ANYmal C with \SI{50}{kg}) and on hardware, including the ANYmal C robot, and (4) detailed evaluation of the control framework and design choices, providing insights for future research in platform-adaptive locomotion control. 

%% file: sections/preliminaries.tex
\section{Preliminaries}
\label{sec:preliminaries}

\subsection{Reinforcement Learning}
The problem is formulated as a time-discrete Markov Decision Process (MDP)~\cite{suttonReinforcementLearningIntroduction1998} in an RL context, with action $\mathbf{a}$ and state $\mathbf{s}$. The agent transitions from state $\mathbf{s_i}$ to $\mathbf{s_{i+1}}$ after taking action $\mathbf{a_i}$, determined by the transition probability $\mathcal{P}(\mathbf{s}_i, \mathbf{a}_i, \mathbf{s}_{i+1})$. The agent receives a reward $\mathrm{r}_i = \mathcal{R}(\mathbf{s}_i, \mathbf{a}_i, \mathbf{s}_{i+1})$ with $\mathcal{R}$ being the reward function. From its initial state $\mathbf{s}_0$ the agent interacts with the environment until it satisfies a terminal condition, such as a finite time horizon or the success/failure of the given task. The cumulative discounted reward collected by the agent is defined as
\begin{equation}
    J\left(\pi\right)\doteq\underset{\mathcal{T}\sim\pi_\theta}{\text{\textup{E}}}\left[\sum_{t=0}^{\infty}{{\gamma}^{t}R\left(\mathbf{s}_{t},\mathbf{a}_{t},\mathbf{s}_{t+1}\right)}\right]\text{,}
    \label{eq:discounted_reward_ex}
\end{equation}
where $\gamma\in\left[0,1\right)$ is the discount factor and $\mathcal{T}$ denotes a trajectory dependent on $\pi_\theta$. The goal of RL is to find the parameters $\theta$ of the policy $\mathrm{\pi}: \mathbf{s} \rightarrow ~a$ that maximizes the expected return defined in Equation~(\ref{eq:discounted_reward_ex}). This work refers to the policy as a function mapping the observation states to the desired control action to achieve similar quadrupedal locomotion across various robots. 

\subsection{System Model}
We model the generalized coordinates $\mathbf{q}$ and velocities $\mathbf{u}$ of any four-legged robot as $\mathbf{q} = \begin{bmatrix}
                            \mathbf{r}_{B} &
                            \mathbf{q}_{B} &
                            \mathbf{q}_{j}
                        \end{bmatrix}$, and $ 
    \mathbf{u} = \begin{bmatrix}
                            \mathbf{v}_{B} &
                            \mathbf{\omega}_{\,B} &
                            \mathbf{\dot{q}}_{j}
                        \end{bmatrix}. $
$\mathbf{r}_B \in \mathbb{R}^3$ is the robot base position, $\mathbf{q}_B\in\mathbb{R}^4$ is the base quaternion with corresponding rotation matrix $\mathbf{R}_{B}\in\mathbb{R}^{3\times3}$. The linear and angular velocities of the base in the world frame are expressed as $\mathbf{v}_{B}\in\mathbb{R}^3$ and $\mathbf{\omega}_{\,B}\in\mathbb{R}^3$. The joint positions are described by $\mathbf{q}_{j}\in\mathbb{R}^{n_{j}}$ and joint velocities $\mathbf{\dot{q}}_{j}\in\mathbb{R}^{n_{j}}$, with $\mathrm{n_j} = 12$ being the total number of actuated joints for any of the robots shown in Figure~\ref{fig:introduction:control_framework}.

%% file: sections/methodology.tex
\section{Methodology}
\label{sec:methodology}

The following section details the robot generation and training method, the architecture chosen for the dynamics encoding network, and the control policy. The range of the kinematic and dynamic robot randomization parameters is provided in Table~\ref{table:appendix:robot_generation_parameters}.

\subsection{Velocity-Tracking Task}
\label{ss:task}
The velocity command is expressed in the base frame as
\begin{equation}
    \mathbf{c}^\text{des} = [\mathrm{v}_x^\text{des}\mathbf{e}_x^B \quad \mathrm{v}_y^\text{des}\mathbf{e}_y^B \quad
    \mathrm{\omega}_z^\text{des}\mathbf{e}_z^B]^T
\end{equation} 
where $\mathrm{v}_x^\text{des}$ and $\mathrm{v}_y^\text{des}$ are the desired heading and lateral velocities expressed along the base frame axis $\mathbf{e}_x^B$ and $\mathbf{e}_y^B$ respectively, and $\mathrm{\omega}_z^\text{des}$ is the desired yaw rate around the vertical axis $\mathbf{e}_z^B$. The velocity command is uniformly sampled within the ranges $v_x^\text{max} = \pm \SI{1}{m/s}$, $v_y^\text{max} = \pm \SI{0.75}{m/s}$, and $\omega_z^\text{max} = \pm \SI{1.5}{rad/s}$. The minimum command duration before resampling is \SI{3}{s} and the maximum is \SI{6}{s}.

\subsection{Robot Generation}
\label{ss:robot_generation}

\subsubsection{Quadruped reference models}
We utilize up to four simplified reference models based on existing quadrupeds to generate a set of 50 robot configurations per model by randomizing their kinematic and dynamic properties. The \textit{reference quadrupeds} used are shown at the bottom of the simulation section on the right of Figure \ref{fig:introduction:control_framework}. The top row shows Unitree's A1~\cite{unitreeroboticsA12023}, Aliengo and Laikago quadrupeds as well as ANYbotics' ANYmal B~\cite{hutterANYmalHighlyMobile2016} and ANYmal C~\cite{ackermanevanANYboticsIntroducesSleek2019}. On the bottom are the simplified versions used as a basis to generate a large set of robots, randomizing their kinematic and dynamic properties. During training, we evaluate single and multiple reference model usage by scaling the chosen reference model and show how successively including more models impacts the robustness and quality of the resulting locomotion policy in Section~\ref{sec:results_and_discussion}.

\subsubsection{Kinematic and dynamic robot model parameter randomization}
Based on the \textit{reference quadrupeds}, we generate 50 variants per robot while randomizing the parameters provided in Table~\ref{table:appendix:robot_generation_parameters} using the RaiSim simulator~\cite{raisimtechRaiSimV1Documentation2023}. 
Suppose a robot can stand collision-free in its nominal configuration for \SI{2}{\second} using the simulator's joint controller. In that case, we admit it into the set of randomized robots used for training. We resample 20\% of this set after $\mathrm{T}_\text{robot}^\text{samp}$. To address the question of how many reference models are necessary, we test the sampling of only one reference model (here with ID 1) as in~\cite{luoMorALLearningMorphologically2024} and show how adding additional models allows for the sharing of learned knowledge of the policy to achieve superior robustness in Figure~\ref{fig:results:robustnessA1andANYmalC}.

\begin{figure}
    \centering
    \includegraphics[width=0.25\textwidth]{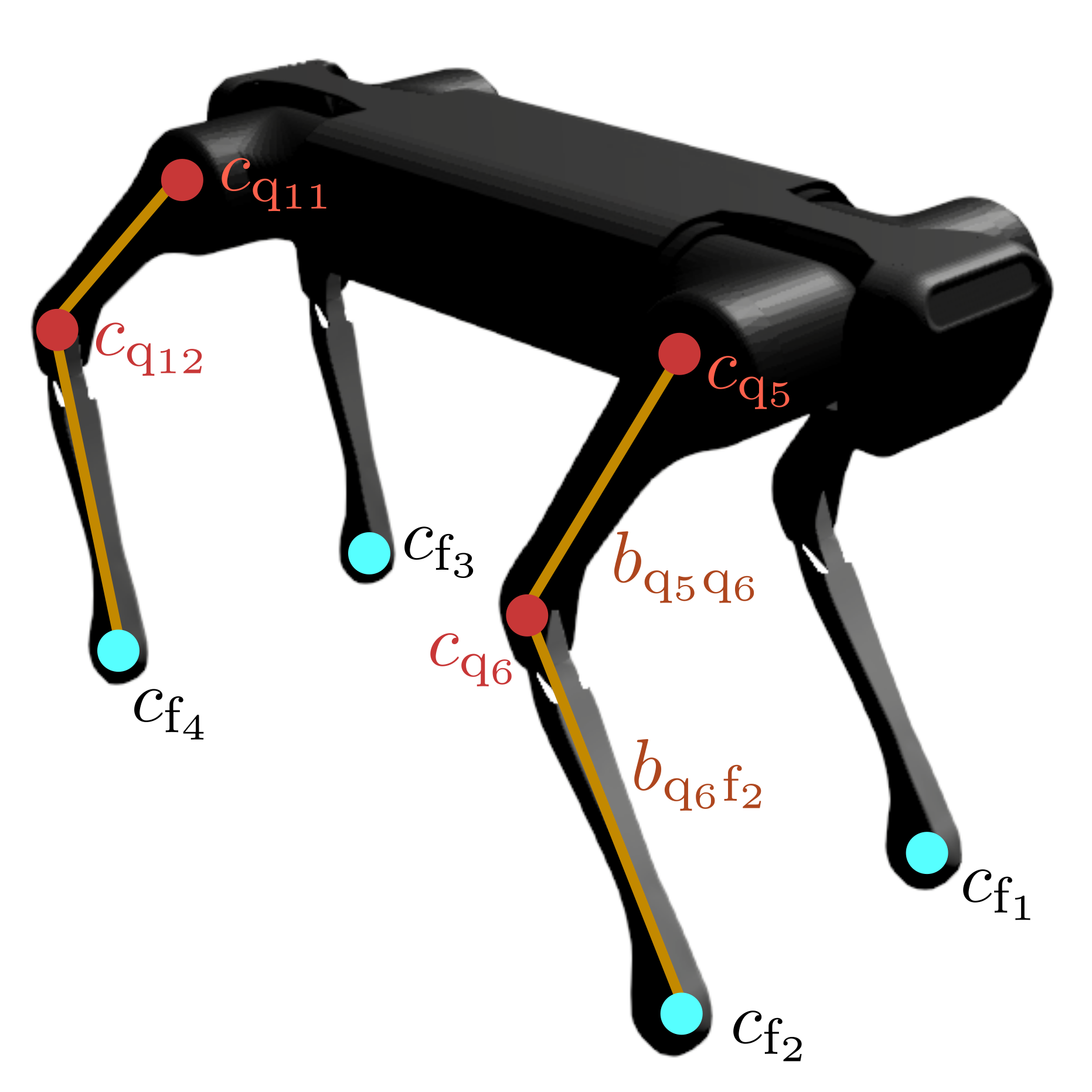}
    \caption{Illustration of the kinematic robot description utilized in this work. The red circles
    represent rotational joints, whereas the blue circles represent fixed joints.}
    \label{fig:preliminaries:robot_description}
        \vspace{-0.5cm}

\end{figure}

We represent the kinematic model of a general quadruped as a combination of individual bodies, $b_{AB}$, linked either via fixed or rotational joints, $A$ and $B$, where $A$ is the parent joint and $B$ is the child joint. An exception to this is the \textit{base} body, represented as $b_{base}$, which is considered to be a floating joint. Each body, $b_{AB}$ has an associated mass, $m_{AB}\in\mathbb{R}$. The position of each child joint, $B$, relative to the parent joint $A$, is represented as $c_B\in\mathbb{R}^3$. As an example, in Fig.~\ref{fig:preliminaries:robot_description}, the body $b_{{\mathrm{q}_5}{\mathrm{q}_6}}$ is linked to joints $\mathrm{q}_5$ and $\mathrm{q}_6$, and has a mass $m_{{\mathrm{q}_5}{\mathrm{q}_6}}$. Additionally, each of the feet has a friction coefficient $\mu_{f}$ associated with it. We also attach a \textit{frame} to each foot with an offset $\mathrm{c}_{f,z}$. The leg configurations are sampled in ranges representing two distinct configurations: A (knees pointing backwards) and X (knees pointing towards the base centre).
    
\begin{table*}
    \small
    \centering
    \begin{tabulary}{\textwidth}{ |C|C|C|C|C| }
    \hline
         \textbf{Parameter(s)} & \multicolumn{4}{c|}{\textbf{Sampling distributions for corresponding base model and ID}} \\ 
         & A1 - 1& Aliengo - 2 & ANYmal B - 4 & ANYmal C - 5\\
        \hline
        
        $c_{\mathrm{q}_1,x}$, $c_{\mathrm{q}_4,x}$, $c_{\mathrm{q}_7,x}$, $c_{\mathrm{q}_{10},x}$ 
        & $\mathcal{U}(0.15, 0.4)$ & $\mathcal{U}(0.15, 0.45)$ & $\mathcal{U}(0.225, 0.45)$ & $\mathcal{U}(0.18, 0.5)$ \\ 
        
        \hline
        
        $c_{\mathrm{q}_1,y}$, $c_{\mathrm{q}_4,y}$, $c_{\mathrm{q}_7,y}$, $c_{\mathrm{q}_{10},y}$ 
        & $\mathcal{U}(0, 0.25)$ & $\mathcal{U}(0., 0.25)$ & $\mathcal{U}(0.05, 0.23)$ & $\mathcal{U}(0.05, 0.27)$ \\ 
        
        \hline
        
        $c_{\mathrm{q}_1,z}$, $c_{\mathrm{q}_4,z}$, $c_{\mathrm{q}_7,z}$, $c_{\mathrm{q}_{10},z}$ 
        & $\mathcal{U}(-0.1, 0.12)$ & $\mathcal{U}(-0.1, 0.12)$ & $\mathcal{U}(-0.18, 0.18)$ & $\mathcal{U}(-0.22, 0.15)$ \\ 
        
        \hline
        
        $c_{\mathrm{q}_2,x}$, $c_{\mathrm{q}_5,x}$, $c_{\mathrm{q}_8,x}$, $c_{\mathrm{q}_{11},x}$ 
        & $\mathcal{U}(-0.1, 0.1)$ & $\mathcal{U}(-0.1, 0.1)$ & $\mathcal{U}(-0.1, 0.15)$ & $\mathcal{U}(-0.1, 0.2)$ \\ 
        
        \hline
        
        $c_{\mathrm{q}_2,y}$, $c_{\mathrm{q}_5,y}$, $c_{\mathrm{q}_8,y}$, $c_{\mathrm{q}_{11},y}$ 
        & $\mathcal{U}(-0.04, 0.13)$ & $\mathcal{U}(0.04, 0.12)$ & $\mathcal{U}(0.015, 0.12)$ & $\mathcal{U}(-0.15, -0.05)$ \\ 
        
        \hline
        
        $c_{\mathrm{q}_2,z}$, $c_{\mathrm{q}_5,z}$, $c_{\mathrm{q}_8,z}$, $c_{\mathrm{q}_{11},z}$ 
        & $\mathcal{U}(-0.1, 0.1)$ & $\mathcal{U}(-0.06, 0.1)$ & $\mathcal{U}(-0.07, 0.07)$ & $\mathcal{U}(-0.1, 0.06)$ \\ 
        
        \hline
        
        $c_{\mathrm{q}_3,x}$, $c_{\mathrm{q}_6,x}$, $c_{\mathrm{q}_9,x}$, $c_{\mathrm{q}_{12},x}$ 
        & $\mathcal{U}(-0.05, 0.18)$ & $\mathcal{U}(-0.05, 0.15)$ & $\mathcal{U}(-0.1, 0.1)$ & $\mathcal{U}(-0.1, 0.2)$ \\ 
        
        \hline
        
        $c_{\mathrm{q}_3,y}$, $c_{\mathrm{q}_6,y}$, $c_{\mathrm{q}_9,y}$, $c_{\mathrm{q}_{12},y}$ 
        & $\mathcal{U}(-0.05, 0.1)$ & $\mathcal{U}(-0.05, 0.1)$ & $\mathcal{U}(-0.05, 0.16)$ & $\mathcal{U}(-0.2, 0.15)$ \\ 
        
        \hline
        
        $c_{\mathrm{q}_3,z}$, $c_{\mathrm{q}_6,z}$, $c_{\mathrm{q}_9,z}$, $c_{\mathrm{q}_{12},z}$ 
        & $\mathcal{U}(-0.24, -0.12)$ & $\mathcal{U}(-0.28, -0.1)$ & $\mathcal{U}(-0.35, -0.18)$ & $\mathcal{U}(-0.35, -0.18)$ \\ 
        
        \hline
        
        $m_{base}$
        & $\mathcal{U}(2, 28)$ & $\mathcal{U}(4, 30)$ & $\mathcal{U}(6, 40)$ & $\mathcal{U}(18, 50)$ \\ 
        
        \hline
        
        $m_{{\mathrm{q}_1}{\mathrm{q}_2}}$, $m_{{\mathrm{q}_4}{\mathrm{q}_5}}$, $m_{{\mathrm{q}_7}{\mathrm{q}_8}}$, $m_{{\mathrm{q}_{10}}{\mathrm{q}_{11}}}$
        & $\mathcal{U}(0.25, 1)$ & $\mathcal{U}(0.25, 2.6)$ & $\mathcal{U}(0.5, 3)$ & $\mathcal{U}(1.4, 4)$ \\ 
        
        \hline
        
        $m_{{\mathrm{q}_2}{\mathrm{q}_3}}$, $m_{{\mathrm{q}_5}{\mathrm{q}_6}}$, $m_{{\mathrm{q}_8}{\mathrm{q}_9}}$, $m_{{\mathrm{q}_{11}}{\mathrm{q}_{12}}}$
        & $\mathcal{U}(0.5, 4)$ & $\mathcal{U}(0.4, 3)$ & $\mathcal{U}(0.6, 4.5)$ & $\mathcal{U}(1.8, 5)$ \\ 
        
        \hline
                        
        $m_{{\mathrm{q}_3}{\mathrm{f}_1}}$, $m_{{\mathrm{q}_6}{\mathrm{f}_2}}$, $m_{{\mathrm{q}_9}{\mathrm{f}_3}}$, $m_{{\mathrm{q}_{12}}{\mathrm{f}_{4}}}$
        & $\mathcal{U}(0.08, 0.9)$ & $\mathcal{U}(0.1, 0.5)$ & $\mathcal{U}(0.15, 0.6)$ & $\mathcal{U}(0.25, 1)$ \\ 
        
        \hline
        
        $c_{\mathrm{f}_1,z}$, $c_{\mathrm{f}_2,z}$, $c_{\mathrm{f}_3,z}$, $c_{\mathrm{f}_{4},z}$ 
        & $\mathcal{U}(-0.025, 0.12)$ & $\mathcal{U}(-0.025, 0.12)$ & $\mathcal{U}(-0.025, 0.12)$ & $\mathcal{U}(-0.025, 0.12)$ \\ 
        
        \hline
        
        $q^n_1$, $q^n_4$, $q^n_7$, $q^n_{10}$
        &  \multicolumn{4}{c|}{$\mathcal{U}(-0.15, 0.15)$} \\ 
        
        \hline
                        
        (Configuration X) $q^n_2$, $q^n_5$
        &  \multicolumn{4}{c|}{$\mathcal{U}(0.3, 0.9)$} \\ 
        
        \hline
        
        (Configuration X) $q^n_8$, $q^n_{11}$
        &  \multicolumn{4}{c|}{$\mathcal{U}(0.3, 0.9)$} \\ 
        
        \hline

        (Configuration X) $q^n_3$, $q^n_6$, $q^n_9$, $q^n_{12}$
        &  \multicolumn{4}{c|}{$\mathcal{U}(-1.2, -0.6)$} \\ 
        
        \hline
                        
        (Configuration A) $q^n_2$, $q^n_5$
        &  \multicolumn{4}{c|}{$\mathcal{U}(0.3, 0.9)$} \\ 
        
        \hline
        
        (Configuration A) $q^n_8$, $q^n_{11}$
        &  \multicolumn{4}{c|}{$\mathcal{U}(0.3, 0.9)$} \\ 
        
        \hline

        (Configuration A) $q^n_3$, $q^n_6$, $q^n_9$, $q^n_{12}$
        &  \multicolumn{4}{c|}{$\mathcal{U}(-1.8, -0.7)$} \\ 
        
        \hline

        $K_p$
        &  $\mathcal{U}(15, 80)$ & $\mathcal{U}(15, 80)$ & $\mathcal{U}(30, 120)$ & $\mathcal{U}(35, 120)$ \\ 
        
        \hline

        $K_d$
        & \multicolumn{4}{c|}{$\mathcal{U}(0.2, 3)$} \\

        \hline

        $\tau_j^{max}$
        &  $\mathcal{U}(15, 120)$ & $\mathcal{U}(15, 50)$ & $\mathcal{U}(40, 80)$ & $\mathcal{U}(40, 140)$ \\ 
        
        \hline

        $\mu_f$
        &  \multicolumn{4}{c|}{$\mathcal{U}(0.2, 1.1)$} \\ 
        
        \hline
    \end{tabulary}
    \caption{Summary of the sampling parameters for the kinematic and dynamic morphology parameters of the quadrupeds.}
    \label{table:appendix:robot_generation_parameters}
\end{table*}

\subsubsection{Joint controller type randomization}
We use a Proportional-Derivative (PD) controller to track the desired joint positions $\mathbf{q}_j^\text{des}$ for the joint with the index $j\in\left\{ 1,\ldots,12 \right\}$ in Figure~\ref{fig:introduction:control_framework}, which outputs a joint torque signal $\tau_j^{\text{cmd}}$ that is applied to the system actuators as 
\begin{equation}
    \tau_j^{\text{cmd}} = K_p(\mathrm{q}^{\text{des}}_j-\mathrm{q}_j)
    - K_d\mathrm{\dot{q}}_j. 
    \label{eq:impedance_controller_simplified}
\end{equation}
$K_p$ and $K_d$ refer to the position and velocity tracking gains, respectively. 


\subsubsection{Actuation command tracking delay modeling}
We introduce actuation command tracking delays in the simulator. Moreover, robotic systems often exhibit complex non-linear actuation dynamics. Therefore, to add this kind of complex dynamics into our simulation, we use actuator networks as in \cite{hwangboLearningAgileDynamic2019} that randomly replace Equation~(\ref{eq:impedance_controller_simplified}) during training. The data used for training the network was collected on the ANYmal B and C hardware.

\subsection{Generalized Morphological Controller}
PAL consists of a dynamics inference module and a control module as described in Figure \ref{fig:introduction:control_framework}. A crucial part is the DIM, which produces a latent dynamics vector. This estimated vector is then integrated into the state space of the locomotion policy, represented by the control module. The CM uses the latent dynamics vector to generate an action vector that specifies the desired joint positions while estimating the base velocity. The robot's low-level PD actuation controllers subsequently track these joint positions. We undertake training based on the model generation method described in Section~\ref{ss:robot_generation}. Each of the robots has been generated using a different set of physical and actuation characteristics to assess the efficacy and adaptability of our proposed method. For our evaluation, we replace the GRU-based DIM with the estimator structure presented by \textit{Luo et al.} in \cite{luoMorALLearningMorphologically2024} and adjust the state spaces for the CM accordingly.

\subsection{Dynamics Encoding}
Motivated by the work of~\cite{zargarbashiMetaLocoUniversalQuadrupedal2024, schmidhuberLearningThinkAlgorithmic2015, liDeepimDeepIterative2018a}, we propose to improve the hidden state $\mathbf{l}_d$ estimate of the quadruped dynamics in the DIM using a recurrent network with hidden state size of $\mathbf{h}\in\mathbb{R}^{36}$ in the form of a GRU. To visually represent our approach, we illustrate the specific cell structure of our recurrent network in Figure \ref{fig:introduction:control_framework} with $h_t$ representing the hidden state. This depiction highlights the intricate architecture we utilize to enhance the estimation of the hidden state $\mathbf{l}_d$ for quadruped dynamics. The network is trained offline using simulation data.

\subsection{State and action spaces}\label{ss:state-and-action-space}
The observation of the locomotion controller consists of the terms defined in Table~\ref{tab:state_definitions}. The estimator input $\mathbf{s}_t^e$ follows the same structure as the actor and critic input $\mathbf{s}_t$. The GRU observation tuple $\mathbf{s}_g$ uses a reduced set of features including the recurrent state $\mathbf{s}_d$.

The observation structures used in the system are: 
\begin{subequations}
\label{eq:method:obs_definitions}
\begin{align}
    \mathbf{s}_t &= \langle \mathbf{s}_R, \mathbf{s}_v, \mathbf{s}_j, \mathbf{s}_\ast, \mathbf{s}_n, \mathbf{s}_c, \mathbf{s}_{h,q}, \mathbf{s}_{h,\dot{q}}, \mathbf{s}_{h,q^\ast}, \mathbf{s}_d \rangle, \label{eq:method:state_loco}\\
    \mathbf{s}_g &= \langle \mathbf{s}_R, \mathbf{s}_v, \mathbf{s}_j, \mathbf{s}_\ast, \mathbf{s}_d \rangle = \langle \mathbf{x}_t, \mathbf{s}_d \rangle, \label{eq:method:state_rnn}\\
    \mathbf{s}_t^e &= \langle \mathbf{s}_R, \tilde{\mathbf{s}}_v, \mathbf{s}_j, \mathbf{s}_\ast, \mathbf{s}_n, \mathbf{s}_c, \mathbf{s}_{h,q}, \mathbf{s}_{h,\dot{q}}, \mathbf{s}_{h,q^\ast}, \mathbf{s}_d \rangle. \label{eq:method:obs_estimator}
\end{align}
\end{subequations}


The action space comprises 12 actions to generate joint position commands that will be tracked using Equation~(\ref{eq:impedance_controller_simplified}). The action is sampled from a Gaussian distribution with zero mean and standard deviation  $\sigma_a = 0.6$ and then added to the nominal joint configuration.
Compared to the MorAL policy of \textit{Luo et al.} \cite{luoMorALLearningMorphologically2024}, we only use the state history of the past two control steps instead of the past five. Additionally, we removed the height map information from the state observations in our implementation of MorAL, as we evaluated our policies presented here on flat terrain only.

\begin{table}[ht]
    \centering
    \centering
    \footnotesize  
    \setlength{\tabcolsep}{4pt}  
    \renewcommand{\arraystretch}{1.0}  
    \begin{tabular}{|c|c|l|}
        \hline
        \textbf{State}      & \textbf{Definition}                                                           & \textbf{Description} \rule{0pt}{2.6ex} \\
        \hline
        $\mathbf{s}_R$      & $\mathbf{e}_z^B$                                                              & Base orientation (roll and pitch) \rule{0pt}{2.6ex} \\
        $\mathbf{s}_v$      & $[(\mathrm{R}_B\mathbf{v}_B)^T \quad (\mathrm{R}_B\mathbf{\omega}_B)^T]^T $   & Base twist \rule{0pt}{2.6ex} \\
        $\tilde{\mathbf{s}}_v$      & $(\mathrm{R}_B\mathbf{\omega}_B)^T $   & Base angular velocity\rule{0pt}{2.6ex} \\
        
        $\mathbf{s}_j$      & $[\mathbf{q}_{j,t} \quad \dot{\mathbf{q}}_{j,t}]^T$                           & Joint position and velocity \rule{0pt}{2.6ex} \\
        $\mathbf{s}_\ast$   & $\mathbf{q}_{j,t}^\ast = \mathbf{q}_{j,t}^\text{des} - \mathbf{q}_j^n$        & Joint position target error \rule{0pt}{2.6ex}\\
        $\mathbf{s}_n$      & $\mathbf{q}_j^\text{n}$                                                              & Joint nominal position  \rule{0pt}{2.6ex}\\
        $\mathbf{s}_c$      & $\mathbf{c}^\text{des}$                                                             & Velocity command \rule{0pt}{2.6ex} \\
        $\mathbf{s}_{h,q}$  & $[\mathbf{q}_{j,t-1} \quad \mathbf{q}_{j,t-2}] ^T$                            & Joint position history \rule{0pt}{2.6ex} \\
        $\mathbf{s}_{h,\dot{q}}$ & $[\dot{\mathbf{q}}_{j,t-1} \quad \dot{\mathbf{q}}_{j,t-2}] ^T$           & Joint velocity history \rule{0pt}{2.6ex} \\
        $\mathbf{s}_{h,q^\ast}$ & $[ {\mathbf{q}_{j,t-1}^\ast} \quad {\mathbf{q}_{j,t-2}^\ast}] ^T$         & Joint position target error history \rule{0pt}{2.6ex} \\     
        $\mathbf{s}_d$      & $\mathbf{l}_d$                                                                & Latent dynamics \rule{0pt}{2.6ex} \\
         \hline 
    \end{tabular}
    \caption{State term definitions.}
    \label{tab:state_definitions}
\end{table}

\subsection{Reward}
The reward is formulated to follow a desired base velocity command as defined in Section \ref{ss:task} with additional rewards added for improved performance. Similar to~\cite{gangapurwalaLearningLowFrequencyMotion2023}, we compute the overall reward function as the total sum of the terms defined in Table~\ref{tab:fp_reward_definitions}. 

\begin{table}[ht!]
    \centering
    \footnotesize  
    \setlength{\tabcolsep}{4pt}  
    \renewcommand{\arraystretch}{1.0}  
    \begin{tabular}{|l|l|}
        \hline
        \textbf{Reward} & \textbf{Description} \\
        \hline
        Base linear velocity & $\mathrm{r}_{v} = 3 \cdot (1-\tanh(4\Vert \mathbf{c}_{xy}^\text{des} - \mathbf{c}_{xy} \Vert^2))$ \\
        \hline
        Base angular velocity & $\mathrm{r}_\omega = 1.75 \cdot (1-\tanh(2\Vert \mathbf{c}_{z}^\text{des} - \mathbf{c}_{z} \Vert^2))$ \\
        \hline
        Base orientation & $\mathrm{r}_R = -5 \cdot \tanh(R_{B})^2$ \\
        \hline
        Base height & $\mathrm{r}_h = -20 \cdot \tanh((r_z - r_n)^2)$ \\
        \hline
        Base undesired motion & $\mathrm{r}_b = -0.5 \cdot (v_z^2 + 0.25 \cdot (|\omega_x| + |\omega_y|))$ \\
        \hline  
        Joint position & $\mathrm{r}_q = -0.2 \cdot ||q_t - q^{nominal}||^2$ \\
        \hline
        Joint velocity & $\mathrm{r}_{\dot{q}} = -3 \times 10^{-4} \cdot \Vert \dot{q}_t \Vert^2$ \\
        \hline
        Joint acceleration & $\mathrm{r}_{\ddot{q}} = -2 \times 10^{-7} \cdot \Vert \ddot{q}_t \Vert^2$ \\
        \hline
        Joint torque & $\mathrm{r}_{\tau} = -3.5 \times 10^{-5} \cdot \Vert \tau \Vert^2$ \\
        \hline
        Joint action smoothness & $\mathrm{r}_{s1} = -0.1 \cdot ||q_t^{des} - q_{t-1}^{des}||^2$ \\
        \hline
        Joint action smoothness 2 & $\mathrm{r}_{s2} = -0.05 \cdot ||q_t^{des} - 2\cdot q_{t-1}^{des} + q_{t-2}^{des}||^2$ \\
        \hline
        Foot slip & $\mathrm{r}_{\mu,i} = -0.15 \cdot c_{f,i} \cdot \Vert v_{f,xy,i} \Vert$ \\
        \hline
        Air time & $\mathrm{r}_{air,i} = -3 \cdot \begin{cases}
            \text{if}~||\mathbf{c}^\text{des}||^2 = 0: & -T_{i,swing} \\
            \text{else:} & T_{i,swing} - 0.5
        \end{cases}$ \\
        \hline 
    \end{tabular}
    \caption{Reward term definitions.}
    \label{tab:fp_reward_definitions}
\end{table}

The index $i\in\left\{1,2,3,4\right\}$ represents the corresponding front left, front right, hind left and hind right leg. $T_{i,stance}$ and $T_{i,swing}$ represent the time since the last touchdown and takeoff, respectively, while being initialized to zero when a transition occurs. $\mathrm{c}_{f,i}$ refers to the contact state of foot $i$. $\mathrm{c}_{f,i}=1$ implies that foot $i$ is in contact with the ground. $R_{B}$ represents the base orientation with respect to the globally fixed world frame. $v_{f,xy,i}$ represents the tangential foot velocity at its contact point. 

All policies are trained with the same reward structure and weights, thus resulting in a similar locomotion gait pattern.

\subsection{Training Setup - RL Algorithm}
\label{sec:methodology:training_setup}

We train the locomotion policy using Proximal Policy Optimization (PPO)~\cite{schulman2017}. The locomotion policy $\pi_\theta$ and the velocity estimator policy are parameterized as a Multi-Layer Perceptron (MLP) with leaky-relu activation. The locomotion policy is modelled with two hidden layers of size 512, and the velocity estimator policy is modelled with two layers of size 512 and 256. The inputs to $\pi_\theta$ are the state observation $\mathbf{s}_t\in\mathbb{R}^{168}$ and the output the control actions $\mathbf{a}_t$. For the estimator, the input is $\mathbf{s}_t^e\in\mathbb{R}^{165}$ with the base linear velocity as output. This formulation of the state-action space is directly adapted from \textit{Gangapurwala et al.} ~\cite{gangapurwalaGuidedConstrainedPolicy2020, gangapurwalaLearningLowFrequencyMotion2023}. For a detailed analysis of the MDP design, we refer the reader to the original work~\cite{hwangboLearningAgileDynamic2019}. The CM and DIM run at a frequency of \SI{100}{\hertz}. Having $t$ represent the current measurements for the states defined in Table \ref{tab:state_definitions}, the history is sampled at $t-1$ and $t-2$ which correspond to a state history sampled at \SI{10}{\milli\second} or \SI{20}{\milli\second} respectively. We apply a penalty of $-1$ for early termination if a collision is detected as a self-collision or ground collision with a body other than a foot.

The training time and hyperparameters were hand-tuned based off~\cite{gangapurwalaLearningLowFrequencyMotion2023} and are provided in Table~\ref{tab:method:hyperparameters}. For the recurrent encoding network, we use a Batched Back-Propagation through Time (BPTT) approach to reduce the computational overhead during policy iteration~\cite{mozerFocusedBackpropagationAlgorithm2013}. Each episode has 600 timesteps. We trained the policies using 30 CPU cores (@2.9 GHz) and an NVIDIA RTX 3090 GPU. The hyperparameters used for the algorithm choice as described in Table~\ref{tab:method:hyperparameters}.

\begin{table}[h!]
    \centering
    \small
    \setlength{\tabcolsep}{3pt}
    \begin{tabular}{|l|c||l|c|}
        \hline
        \textbf{Parameter} & \textbf{Value} & \textbf{Parameter} & \textbf{Value} \\
        \hline
        Discount factor, $\gamma$ & 0.9962 & Entropy coefficient & 0 \\
        Learning rate & adaptive & Value coefficient & 0.5 \\
        Batch size & 63000 & GAE & True \\
        Mini-batch size & 8 & GAE $\lambda$ & 0.95 \\
        Epochs & 4 & Steps per iteration & 140 \\
        Parallel envs, $n_{env}$ & 450 & PPO time/iteration & $\sim$1.5 s \\
        BPTT Batch Size & 50 & Total training time & $\sim$20 h \\
        DIM training time & $\sim$6 h & Robot resample time & 25 s \\
        \hline 
    \end{tabular}
    \caption{RL Training time and hyperparameters.}
    \label{tab:method:hyperparameters}
\end{table}

%% file: sections/results_and_discussion.tex
\section{Simulation and hardware validation}
\label{sec:results_and_discussion}

In this section, we compare the effectiveness of our controllers on the locomotion task for quadrupedal robots with different kinematic and dynamic parameters. 

\subsection{Performance Benchmark}
The comparative evaluation is performed among the following reference-free locomotion controllers for the flat terrain walking task:
\begin{enumerate}
    \item \textbf{GenLoco}~\cite{fengGenLocoGeneralizedLocomotion2022}: This controller takes a long state history as input without explicitly conditioning the locomotion policy on estimating robot-specific kinematic or dynamic parameters. 
    \item \textbf{MorAL}~\cite{luoMorALLearningMorphologically2024}: The policy is trained concurrently with an MLP that explicitly estimates the body states and kinematic morphology parameters. We remove terrain information for this work as the evaluation occurs on flat terrain only and adjust the critic observations accordingly. 
    \item \textbf{PAL (Ours)}: The policy is trained with a GRU-based estimator for dynamic parameter estimation, described in Section~\ref{sec:methodology}.
\end{enumerate}
We also train a base velocity estimator concurrently for each controller structure.

Additionally, we compare the effects of including different sets of reference quadrupeds in the training sets of viable configurations. For the MorAL~\cite{luoMorALLearningMorphologically2024} baseline we include only one robot model (the Unitree A1 here with ID $\left\{1\right\}$ in Figure~\ref{fig:introduction:control_framework}) and increase the number of reference robots that we expose to the policy during training to include the robot IDs $\left\{1, 4\right\}$ or $\left\{1, 2, 4, 5\right\}$. Note that the specific quadrupeds were never included in any of the training sets, but instead used as a base model in which we change the sampled kinematic and dynamic parameters described in Table~\ref{table:appendix:robot_generation_parameters}.

\begin{figure}
    \centering
    \includegraphics[width=0.495\textwidth]{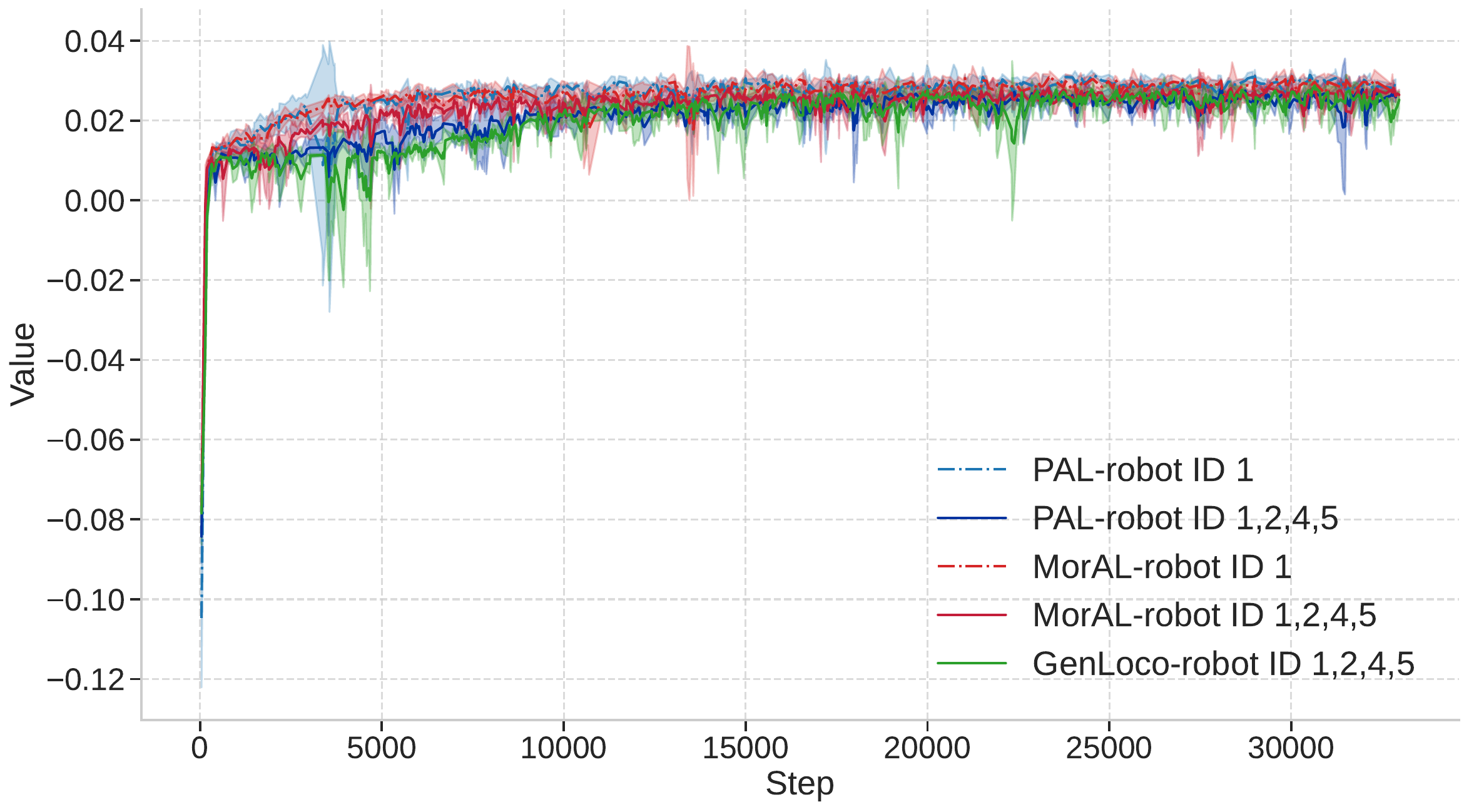}
    \caption{The average learning performance of three controller architectures trained with different robot ID sets. The results are obtained from three different random seeds each.}
    \vspace{-0.5cm}
    \label{fig:results:rewards}
\end{figure}

Figure~\ref{fig:results:rewards} shows that the learning performance achieves similar results regarding final average reward. When using only robot ID $\left\{1\right\}$ as in \cite{luoMorALLearningMorphologically2024} during training, the policy learns faster for all controller architectures. The final reward is similar for any number of robot configurations used for the training set when comparing the robot training ID set $\left\{1,2,4,5\right\}$ which is in line with findings of~\cite{bohlingerOnePolicyRun2025}. We could not safely deploy our implementation of the GenLoco controller on hardware. As a result, further investigation in this section only happens on the PAL and MorAL controller architectures with different sets of robot IDs during training.  

\subsection{Robustness}
We use the definition of success rate (SR) as a performance metric for robustness as defined by~\textit{Gangapurwala et al.}~\cite{gangapurwalaLearningLowFrequencyMotion2023}:
\begin{equation}
    \text{SR} = 1 - \frac{N_e}{N_T}.
\end{equation}
With $N_e$ referring to the number of rollouts that terminated early due to a prohibited collision and $N_T$ being the total number of rollouts. Using $N_T=100$, we randomize the base linear velocity command with $0.75 \cdot v_\text{x,y}^\text{max}$ from Section~\ref{ss:task} for \SI{4}{\second} before resampling. The same early termination criteria defined in Section~\ref{ss:robot_generation} are applied. We test four parameter changes, measuring each time the SR: horizontal force perturbation onto the base, changes in foot contact friction, actuation latency, and base mass. 

\begin{figure*}
    \centering
    \includegraphics[width=0.9\textwidth]{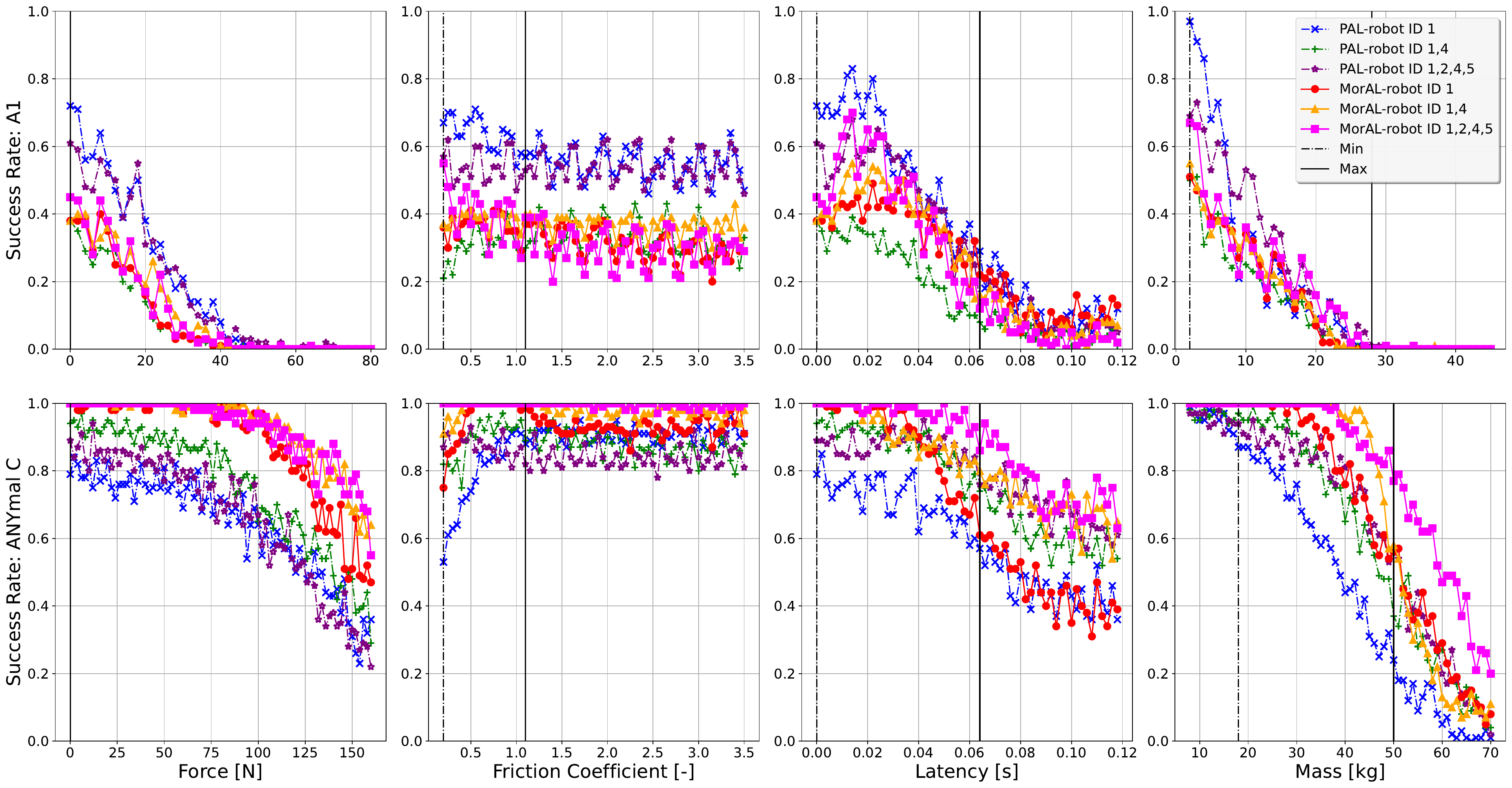}
    \caption{Success rate observed for various motion control policies for different perturbations and dynamics parameters for the A1 (top row) and ANYmal C (bottom row) quadrupeds. We measure the success rate over a wide range of parameters, showing in black the minimum and maximum training range of the disturbance parameters for base perturbation in the horizontal plane during walking, friction coefficient, latency, and base mass change. For the smaller robots, such as the A1, the PAL controller performs best, with less influence on the robot ID set choice during training (top row). For larger quadrupeds such as the ANYmal C (bottom row), we achieve best performance with policies exposed to larger robot ID sets during training. The MorAL architecture with robot ID set of $\left\{1,2,4,5\right\}$ results in the most robust locomotion performance.}
    \label{fig:results:robustnessA1andANYmalC}
\end{figure*}

A PAL latent embedding presented in the DIM performs better for the smaller robot A1 in all test cases, as presented in the top row of Figure~\ref{fig:results:robustnessA1andANYmalC}. By conditioning the locomotion policy on explicitly estimating the kinematic morphology robot parameters with the MorAL architecture, we achieve the best results for larger quadrupedal locomotion as seen on the bottom row. Note that by including ANYmal B into the robot reference ID set of $\left\{1,4\right\}$ improves the robustness for both PAL and MorAL controllers, though the best results are with larger reference ID sets such as $\left\{1,2,4,5\right\}$. We attribute this to the exposure of the policy of the more realistic weight distribution of the quadrupeds in training compared to the evaluated robot. This shows the difficulty of uniformly sampling robot kinematic and dynamic parameters using a single quadruped base model during training time.

Therefore, by introducing additional information related to the system characteristics in the state space, the agent can condition its behaviour based on the dynamics, ensuring a successful and robust deployment of the policy at runtime across a range of previously unseen quadrupeds. The accompanying video demonstrates zero-shot transfer onto the full range of robot IDs.

\subsection{Sim-to-Real Transfer}
\label{ss:results:sim_2_real_transfer}

\begin{figure}
    \centering
    \includegraphics[width=0.495\textwidth]{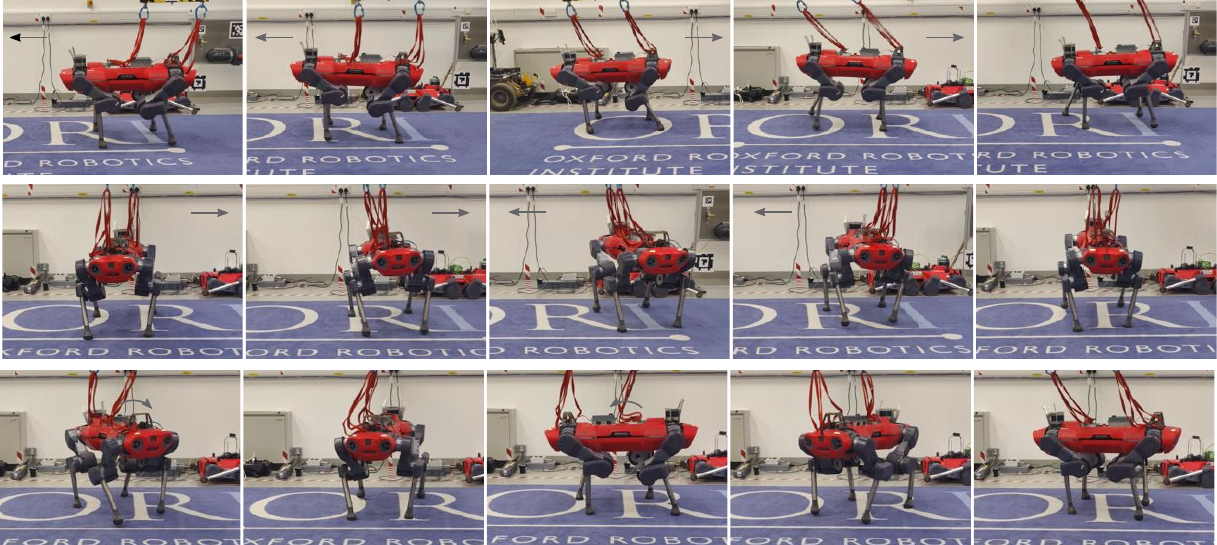}
    \caption{PAL locomotion policy deployed on real ANYmal C robot. Heading (top), lateral (middle), and turn response (bottom).}
    \label{fig:results:anymal_hardware_results}
\end{figure}

In our preliminary experiments, we transferred the trained dynamics encoding and locomotion policy networks to the ANYmal C quadruped. Using the same actuation tracking gains tested in simulation ($K_p=85$ and $K_d=5$) resulted in extremely aggressive behaviour on the hardware compared to smooth simulation behaviour. Following the introduction of latency into the policy training, we successfully transferred the trained networks to the real robot ANYmal C, as depicted in Figure~\ref{fig:results:anymal_hardware_results}. We observe that the tracking behaviour on the real quadruped closely resembles the performance of the simulated one. 

\subsection{Hardware Velocity Command Tracking}
Finally, we evaluate the effectiveness of the learned controllers on the robot in the real world. We successfully deploy PAL and MorAL controller policies with different ranges of quadruped robot IDs. The results in Table~\ref{tab:pal_and_moral_velocityTracking} show the Root Mean Square Error (RMSE) for command tracking by comparing the target velocity command with the estimated base linear heading, lateral, and turn velocity. 

\begin{table}[h!]
    \centering
    \begin{tabular}{|l|l|c|c|c|}
    \hline
        Controller & ID & RMSE-X & RMSE-Y & RMSE-$\theta$ \\
        \hline
        PAL   & 1    & 0.2193 & 0.1202 & 0.2729 \\
        PAL   & 1,4   & 0.2019 & 0.0994 & 0.2734 \\
        PAL   & 1,2,4,5 & 0.1652 & 0.1173 & 0.2713 \\
        \hline
        MorAL & 1    & 0.1339 & 0.0836 & 0.2664 \\
        MorAL & 1,4   & 0.1243 & 0.0970 & 0.2122 \\
        MorAL & 1,2,4,5 & \textbf{0.0885} & \textbf{0.0772} & \textbf{0.1877} \\
        \hline
    \end{tabular}
    \caption{Walking test: velocity command tracking error comparison with RMSE of heading (-X), lateral (-Y), and turn (-$\theta$) commands on the hardware ANYmal C.}
    \label{tab:pal_and_moral_velocityTracking}
\end{table}

The hardware results demonstrate the improvement of using multiple quadrupeds to achieve the best-performing locomotion policy. Choosing a distinct set of robots during training allows for the lowest RMSE. We find that the MorAL architecture achieves more stable locomotion on ANYmal C, which is in line with our simulation-based robustness test results in Figure~\ref{fig:results:robustnessA1andANYmalC}. The MorAL policy exposed during training to the robot ID set $\left\{1,2,4,5\right\}$, reduces the tracking error in heading and turn direction by approximately 30\% compared to the baseline. We observed that PAL trained with fewer robot models, such as using the robot ID $\left\{1\right\}$, sometimes tends to tilt over but can catch itself during locomotion. We refer to the supplementary video material for a visual example. We did not observe such behavior for MorAL controllers and attribute this to recurrent neural networks' tendency to overfit simulation dynamics~\cite{siekmannLearningMemoryBasedControl2020}.  

\subsection{Hardware Base Velocity Estimator}
The velocity estimate is another measure of the quality of the controllers and robot IDs chosen. We compute the RMSE of the trained MLP estimator network with the onboard manufacturer ANYmal C state estimator in Table \ref{tab:pal_and_moral_baseEstimator}. Training our controllers with robot IDs $\left\{1,2,4,5\right\}$ achieves an improvement of around 40\% over controllers trained with only robot reference ID $\left\{1\right\}$. The choice of architecture appears of equal quality, indicating that both morphology parameter conditioning and latent dynamics embedding contain relevant information for successful base velocity estimation. We conclude that future research could combine both architectures to further the quality of the estimator output. The results demonstrate the importance of choosing a suitable set of quadrupeds for high-quality hardware estimates during deployment.

\begin{table}[h!]
\centering
    \begin{tabular}{|l|l|c|c|c|}
        \hline
        Controller & ID & RMSE-X & RMSE-Y & RMSE-Z \\
        \hline
        PAL   & 1    & 0.1279 & 0.1370 & 0.0750 \\
        PAL   & 1,4   & 0.1295 & 0.1272 & 0.0726 \\
        PAL   & 1,2,4,5 & \textbf{0.1058} & 0.1268 & 0.0738 \\
        \hline
        MorAL & 1    & 0.1712 & 0.2042 & 0.1023 \\
        MorAL & 1,4   & 0.1330 & 0.1351 & 0.0900 \\
        MorAL & 1,2,4,5 & 0.1284 & \textbf{0.1160} & \textbf{0.0560} \\
        \hline
    \end{tabular}
    \caption{Walking test: base velocity estimate RMSE of heading (-X), lateral (-Y), and height (-z) estimate with onboard state estimator base twist for hardware ANYmal C.}
    \label{tab:pal_and_moral_baseEstimator}
\end{table}

%% file: sections/conclusion_and_future_work.tex
\section{Conclusion and Future Work}
\label{sec:conclusion_and_future_work}

We presented \textbf{P}latform \textbf{A}daptive \textbf{L}ocomotion (PAL), a policy that adapts to diverse quadrupedal robots by conditioning on temporally local system dynamics. Our results demonstrate that dynamics encoding and randomization techniques alone can be sufficient for successful hardware transfer of a \textit{universal} locomotion policy.

While real-world deployment on ANYmal C did not reveal performance improvements over a strong reference-free baseline for velocity tracking, PAL achieved higher success rates in simulation on smaller quadrupeds such as the A1 for perturbations in the form of external forces, friction variation, and base mass changes. This highlights PAL's potential for improved robustness in more varied and dynamic environments. Both the baseline approach and PAL achieved similar base velocity estimation performance. We also demonstrate that incorporating a broader set of robot reference models during training significantly enhances policy generalization, reducing the velocity command tracking error by up to 30\% and base velocity estimation error by 40\%. Furthermore, we show that agents benefit from directly inferring system dynamics, suggesting the value of embedding structure over purely temporal encodings such as GRUs.

In conclusion, we suggest that embedding physical priors and careful modeling of robot references during training are key to scaling \textit{universal} locomotion controllers to real-world systems. Future work will explore terrain-perceptive controllers within this context to enable rough terrain locomotion.